\documentclass[10pt,twocolumn,letterpaper]{article}

\usepackage{iccv}
\usepackage{times}
\usepackage{epsfig}
\usepackage{graphicx}
\usepackage{amsmath}
\usepackage{amssymb}

\usepackage{tabulary,multirow,overpic,xcolor}
\usepackage[caption=false]{subfig}

\newcommand{\bd}[1]{\textbf{#1}}
\newcolumntype{x}[1]{>{\centering\arraybackslash}p{#1pt}}
\newcommand{\tablestyle}[2]{\setlength{\tabcolsep}{#1}\renewcommand{\arraystretch}{#2}\centering\footnotesize}
\makeatletter\renewcommand\paragraph{\@startsection{paragraph}{4}{\z@}
	{.5em \@plus1ex \@minus.2ex}{-.5em}{\normalfont\normalsize\bfseries}}\makeatother

\usepackage[breaklinks=true,bookmarks=false]{hyperref}

\iccvfinalcopy 

\def\iccvPaperID{} 
\def\httilde{\mbox{\tt\raisebox{-.5ex}{\symbol{126}}}}

\ificcvfinal\fi

\begin{document}

\title{\vspace{-.5em} Video Transformer Network \\ \vspace{-.5em}}

\author{Daniel Neimark
	\quad Omri Bar
	\quad Maya Zohar
	\quad Dotan Asselmann \vspace{.8em}\\
	Theator\\
	{\tt\small \{danieln, omri, maya, dotan\}@theator.io}
}

\maketitle
\ificcvfinal\thispagestyle{empty}\fi

\begin{abstract}
\vspace{-.4em}
This paper presents VTN, a transformer-based framework for video recognition. Inspired by recent developments in vision transformers, we ditch the standard approach in video action recognition that relies on 3D ConvNets and introduce a method that classifies actions by attending to the entire video sequence information. Our approach is generic and builds on top of any given 2D spatial network. In terms of wall runtime, it trains $16.1\times$ faster and runs $5.1\times$ faster during inference while maintaining competitive accuracy compared to other state-of-the-art methods. It enables whole video analysis, via a single end-to-end pass, while requiring $1.5\times$ fewer GFLOPs. We report competitive results on Kinetics-400 and Moments in Time benchmarks and present an ablation study of VTN properties and the trade-off between accuracy and inference speed. We hope our approach will serve as a new baseline and start a fresh line of research in the video recognition domain. Code and models are available at: \url{https://github.com/bomri/SlowFast/blob/master/projects/vtn/README.md}.
\end{abstract}

\section{Introduction}

\begin{figure}[t]
	\begin{center}
	\includegraphics[width=0.99\linewidth]{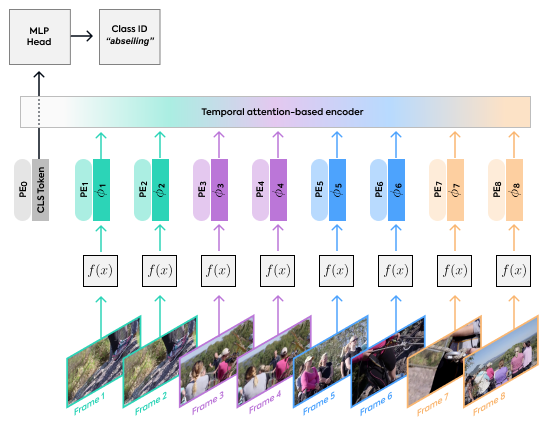}
	\end{center}
	\caption{Video Transformer Network architecture. Connecting three modules: A 2D spatial backbone ($f(x)$), used for feature extraction. Followed by a temporal attention-based encoder (Longformer in this work), that uses the feature vectors ($\phi_i$) combined with a position encoding. The $[CLS]$ token is processed by a classification MLP head to get the final class prediction.}
	\label{fig:arch}
	\vspace{-8pt}
\end{figure}

Attention matters. For almost a decade, ConvNets have ruled the computer vision field \cite{krizhevsky2012imagenet,deng2009imagenet}. 
Applying deep ConvNets produced state-of-the-art results in many visual recognition tasks, i.e., image classification \cite{Simonyan15,he2016deep,tan2019efficientnet}, object detection \cite{girshick2014rich,girshick2015fast,ren2016faster}, semantic segmentation \cite{long2015fully}, object instance segmentation \cite{he2017mask}, face recognition \cite{taigman2014deepface,schroff2015facenet} and video action recognition \cite{donahue2015long,wang2016temporal,carreira2017quo,wang2018non,feichtenhofer2019slowfast,feichtenhofer2020x3d}. But, recently this domination is starting to crack as transformer-based models are showing promising results in many of these tasks \cite{dosovitskiy2021an,carion2020end,touvron2020training,wang2020end,zhou2018end,girdhar2019video}.

Video recognition tasks also rely heavily on ConvNets. In order to handle the temporal dimension, the fundamental approach is to use 3D ConvNets \cite{christoph2016spatiotemporal,carreira2017quo,chao2018rethinking}. In contrast to other studies that add the temporal dimension straight from the input clip level, we aim to move apart from 3D networks. We use state-of-the-art 2D architectures to learn the spatial feature representations and add the temporal information later in the data flow by using attention mechanisms on top of the resulting features. Our approach input only RGB video frames and without any \textit{bells and whistles} (\eg, optical flow, streams lateral connections, multi-scale inference, multi-view inference, longer clips fine-tuning, etc.) achieves comparable results to other state-of-the-art models.

\begin{figure*}[]
	\begin{center}
		\includegraphics[width=0.99\linewidth]{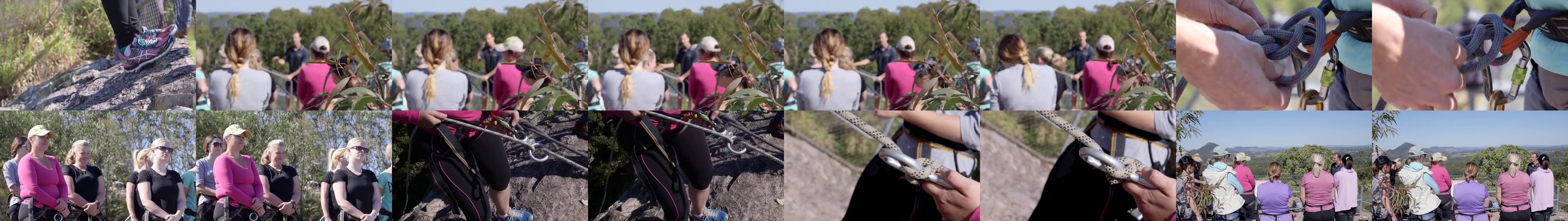}
	\end{center}
	\caption{Extracting 16 frames evenly from a video of the \textit{abseiling} category in the Kinetics-400 dataset \cite{kay2017kinetics}. Analyzing the video’s full context and attending to the relevant parts is much more intuitive than analyzing several clips built around specific frames, as many of these frames might lead to false predictions.}
	\label{fig:frames}
\end{figure*}

Video recognition is a perfect candidate for Transformers. Similar to language modeling, in which the input words or characters are represented as a sequence of tokens \cite{NIPS2017_3f5ee243}, videos are represented as a sequence of images (frames). However, this similarity is also a limitation when it comes to processing long sequences. Like long documents, long videos are hard to process. Even a 10 seconds video, such as those in the Kinetics-400 benchmark \cite{kay2017kinetics}, are processed in recent studies as short, $\httilde2$ seconds, clips.

But how does this clip-based inference would work on much longer videos (i.e., movie films, sports events, or surgical procedures)? It seems counterintuitive that the information in a video of hours, or even a few minutes, can be grasped using only a snippet clip of a few seconds. Nevertheless, current networks are not designed to share long-term information across the full video.

VTN’s temporal processing component is based on a Longformer \cite{beltagy2020longformer}. This type of transformer-based model can process a long sequence of thousands of tokens. The attention mechanism proposed by the Longformer makes it feasible to go beyond short clip processing and maintain global attention, which attends to all tokens in the input sequence. 

In addition to long sequence processing, we also explore an important trade-off in machine learning -- \textit{speed vs. accuracy}. Our framework demonstrates a superior balance of this trade-off, both during training and also at inference time. In training, even though wall runtime per epoch is either equal or greater, compared to other networks, our approach requires much fewer passes of the training dataset to reach its maximum performance; end-to-end, compared to state-or-the-art networks, this results in a $16.1\times$ faster training. At inference time, our approach can handle both multi-view and full video analysis while maintaining similar accuracy. In contrast, other networks’ performance significantly decreases when analyzing the full video in a single pass. In terms of \textit{GFLOPS x Views}, their inference cost is considerably higher than those of VTN, which concludes to a $1.5\times$ fewer GFLOPS and a $5.1\times$ faster validation wall runtime.

Our framework’s structure components are modular (Fig.~\ref{fig:arch}). First, the 2D spatial backbone can be replaced with any given network. The attention-based module can stack up more layers, more heads or can be set to a different Transformers model that can process long sequences. Finally, the classification head can be modified to facilitate different video-based tasks, like temporal action localization.


\section{Related Work}

\paragraph{Spatial-temporal networks.} Most recent studies in video recognition suggested architectures that are based on 3D ConvNets \cite{ji20123d,tran2015learning}. In \cite{christoph2016spatiotemporal}, a two-stream architecture was used, one stream for RGB inputs and another for Optical Flow (OF) inputs. Residual connections are inserted into the two-stream architecture to allow a direct link between RGB and OF layers. The idea of inflating 2D ConvNets into their 3D counterpart (I3D) was introduced in \cite{carreira2017quo}. I3D takes 2D ConvNets and expands its layers into 3D. Therefore it allows to leverage pre-trained state-of-the-art image recognition architectures in the spatial-temporal domain and apply them for video-based tasks. 

Non-local Neural Networks (NLN) \cite{wang2018non} introduced a non-local operation, a type of self-attention, that computes responses based on relationships between different locations in the input signal. NLN demonstrated that the core attention mechanism in Transformers can produce good results on video tasks, however it is confined to processing only short clips. In order to extract long temporal context, \cite{wu2019long} introduced a long-term feature bank that acts as the entire video memory and a Feature Bank Operator (FBO) that computes interactions between short-term and long-term features. However, it requires precomputed features, and it is not efficient enough to support end-to-end training of the feature extraction backbone.

SlowFast \cite{feichtenhofer2019slowfast} explored a network architecture that operates in two pathways and different frame rates. Lateral connections fuse the information between the slow pathway, focused on the spatial information, and the fast pathway focused on temporal information.

The X3D study \cite{feichtenhofer2020x3d} builds on top of SlowFast. It argues that in contrast to image classification architectures, which have been developed via a rigorous evolution, the video architectures have not been explored in detail, and historically are based on expanding image-based networks to fit the temporal domain. X3D introduces a set of networks that progressively expand in different axes, e.g., temporal, frame rate, spatial, width, bottleneck width, and depth. Compared to SlowFast, it offers a lightweight network (in terms of GFLOPS and parameters) with similar performance.

\paragraph{Transformers in computer vision.} The Transformers architecture \cite{NIPS2017_3f5ee243} reached state-of-the-art results in many NLP tasks, making it the de-facto standard. Recently, Transformers are starting to disrupt the field of computer vision, which traditionally depends on deep ConvNets. Studies like ViT and DeiT for image classification \cite{dosovitskiy2021an,touvron2020training}, DETR for object detection and panoptic segmentation \cite{carion2020end}, and VisTR for video instance segmentation \cite{wang2020end} are some examples showing promising results when using Transformers in the computer vision field. Binding these results with the sequential nature of video makes it a perfect match for Transformers.

\paragraph{Applying Transformers on long sequences.} BERT \cite{devlin-etal-2019-bert} and its optimized version RoBERTa \cite{liu2019roberta} are transformer-based language representation models. They are pre-trained on large unlabeled text and later fine-tuned on a given target task. With minimal modification, they achieve state-of-the-art results on a variety of NLP tasks.

One significant limitation of these models, and Transformers in general, is their ability to process long sequences. This is due to the self-attention operation, which has a complexity of $O(n^2)$ per layer ($n$ is sequence length)~\cite{NIPS2017_3f5ee243}.

Longformer \cite{beltagy2020longformer} addresses this problem and enables lengthy document processing by introducing an attention mechanism with a complexity of $O(n)$. This attention mechanism combines a local-context self-attention, performed by a sliding window, and task-specific global attention. 

Similar to ConvNets, stacking up multiple windowed attention layers results in a larger receptive field. This property of Longformer gives it the ability to integrate information across the entire sequence. The global attention part focuses on pre-selected tokens (like the $[CLS]$ token) and can attend to all other tokens across the input sequence.


\begin{table}
	\tablestyle{8.5pt}{1.05}
	\begin{center}
		\begin{tabular}{l|c|c|c}
			\multicolumn{1}{c|}{model} & \multicolumn{1}{c|}{\begin{tabular}[c]{@{}c@{}}pretrain\\ (ImageNet accuracy (\%))\end{tabular}} & \multicolumn{1}{c|}{top-1} & top-5 \\
			\hline
			R50-VTN     & ImageNet (76.2 \cite{paszke2017automatic})         & 71.2  & 90.0  \\
			R101-VTN    & ImageNet (77.4 \cite{paszke2017automatic})         & 72.1  & 90.3  \\
			DeiT-B-VTN  & ImageNet (81.8 \cite{touvron2020training})         & 75.5  & 92.2  \\
			DeiT-BD-VTN & ImageNet (83.4 \cite{touvron2020training})      & 75.6  & 92.4  \\
			ViT-B-VTN   & ImageNet-21K (84.0 \cite{dosovitskiy2021an})     & 78.6  & 93.7 \\
			ViT-B-VTN$^\dagger$   & ImageNet-21K (84.0 \cite{dosovitskiy2021an})     & \textbf{79.8}  & \textbf{94.2} \\
		\end{tabular}
	\end{center}
	\caption{VTN performance on Kinetics-400 validation set for different backbone variations. A full video inference is used. We show top-1 and top-5 accuracy. We report what pre-training was done for each backbone and the related single-crop top-1 accuracy on ImageNet. ($\dagger$) Training with extensive data augmentation.}
	\label{tb:1}
\end{table}

\begin{table*}[t]\centering
	\subfloat[\textbf{Depth}: Comparing different numbers of attention layers in the Longformer. \label{tb:2}]{
		\tablestyle{1.5pt}{1.05}
		\begin{tabular}{c|x{22}x{22}}
			\# layers & top-1 & top-5 \\
			\hline
			1                   & \bd{78.6} & 93.4 \\
			3                   & \bd{78.6} & \bd{93.7} \\
			6                   & 78.5 & 93.6 \\
			12                  & 78.3 & 93.3 \\
			\multicolumn{3}{c}{~}\\
			\multicolumn{3}{c}{~}\\
	\end{tabular}}\hspace{3mm}
	\subfloat[\textbf{Positional embedding}: Evaluating the impact of different types of PE methods, with and without shuffling the input frames. We conducted this experiment with an older experimental setup using a different learning rate scheduler, so the results of the \textit{learned} without \textit{shuffle} are slightly different from what we report in other tables: 78.4\% \textit{vs.} 78.6\%. \label{tb:3}]{
		\tablestyle{6.5pt}{1.05}
		\begin{tabular}{c|x{22}|x{22}x{22}}
			PE  & shuffle & top-1 & top-5 \\
			\hline
			learned & -       & 78.4  & 93.5  \\
			learned & \checkmark       & 78.8  & 93.6  \\
			\hline
			fixed & -       & 78.3  & 93.7  \\
			fixed & \checkmark       & 78.5     & 93.7     \\
			\hline
			no  & -       & 78.6  & 93.7  \\
			no  & \checkmark       & \bd{78.9}  & 93.7   \\	
	\end{tabular}}\hspace{3mm}
	\subfloat[\textbf{Temporal footprint}: Comparing the impact of temporal footprint size (in seconds) and the number of frames in a clip. \label{tb:4}]{
		\tablestyle{1.5pt}{1.05}
		\begin{tabular}{c|c|x{22}x{22}x{22}}			
			\multicolumn{1}{c|}{\begin{tabular}[c]{@{}c@{}}temporal footprint\end{tabular}} & \multicolumn{1}{c|}{\begin{tabular}[c]{@{}c@{}}\# frames\end{tabular}} & top-1 & top-5 \\
			\hline
			2.56               & 16                     & 78.2  & 93.4  \\
			2.56               & 32                     & 78.2     & \bd{93.6}     \\
			5.12               & 16                     & \bd{78.6}  & 93.4  \\
			5.12               & 32                     & 78.5     & 93.5    \\
			10.0               & 16                     & 78.0     & 93.3    \\
			\multicolumn{3}{c}{~}\\
	\end{tabular}}\hspace{3mm}
	\subfloat[\textbf{Finetune}: Training a ViT-B-VTN with three attention layers with and without fine-tuning the 2D backbone. All other hyperparameters remain the same.  \label{tb:5}]{
		\tablestyle{2pt}{1.05}
		\begin{tabular}{c|x{22}x{22}}
			finetune & top-1 & top-5 \\
			\hline
			-                & 71.6  & 90.3  \\
			\checkmark         & \bd{78.6}  & \bd{93.7}  \\
			\multicolumn{3}{c}{~}\\
			\multicolumn{3}{c}{~}\\
			\multicolumn{3}{c}{~}\\
			\multicolumn{3}{c}{~}\\
	\end{tabular}}\hspace{3mm}
	\vspace{.25em}
	\caption{Ablation experiments on Kinetics-400. The results are top-1 and top-5 accuracy (\%) on the validation set using the full video inference approach.}
	\label{tb:main2}
\end{table*}

\section{Video Transformer Network}

Video Transformer Network (VTN) is a generic framework for video recognition. It operates with a single stream of data, from the frames level up to the objective task head. In the scope of this study, we demonstrate our approach using the action recognition task by classifying an input video to the correct action category.

The architecture of VTN is modular and composed of three consecutive parts. A 2D spatial feature extraction model (spatial backbone), a temporal attention-based encoder, and a classification MLP head. Fig.~\ref{fig:arch} demonstrates our architecture layout.

VTN is scalable in terms of video length during inference, and enables the processing of very long sequences. Due to memory limitation, we suggest several types of inference methods. (1) Processing the entire video in an end-to-end manner. (2) Processing the video frames in chunks, extracting features first, and then applying them to the temporal attention-based encoder. (3) Extracting all frames’ features in advance and then feed them to the temporal encoder.

\subsection{Spatial backbone}
\label{section:spatial_backbone}

The spatial backbone operates as a learned feature extraction module. It can be any network that works on 2D images, either deep or shallow, pre-trained or not, convolutional- or transformers-based. And its weights can be fixed (pre-trained) or trained during the learning process.

\subsection{Temporal attention-based encoder}
\label{section:temporal}

As suggested by \cite{NIPS2017_3f5ee243}, we use a Transformer model architecture that applies attention mechanisms to make global dependencies in a sequence data. However, Transformers are limited by the number of tokens they can process at the same time. This limits their ability to process long inputs, such as videos, and incorporate connections between distant information.

In this work, we propose to process the entire video at once during inference. We use an efficient variant of self-attention, that is not all-pairwise, called Longformer \cite{beltagy2020longformer}. Longformer operates using sliding window attention that enables a linear computation complexity. The sequence of feature vectors of dimension $d_\text{backbone}$ (Sec.~\ref{section:spatial_backbone}) is fed to the Longformer encoder. These vectors act as the 1D tokens embedding in the standard Transformer setup.

Like in BERT \cite{devlin-etal-2019-bert} we add a special classification token ($[CLS]$) in front of the features sequence. After propagating the sequence through the Longformer layers, we use the final state of the features related to this classification token as the final representation of the video and apply it to the given classification task head. Longformer also maintains global attention on that special $[CLS]$ token.

\subsection{Classification MLP head}
Similar to \cite{dosovitskiy2021an}, the classification token (Sec.~\ref{section:temporal}) is processed with an MLP head to provide a final predicted category. The MLP head contains two linear layers with a GELU non-linearity and Dropout between them. The input token representation is first processed with a Layer normalization.

\subsection{Looking beyond a short clip context}
\label{section:looking}

The common approach in recent studies for video action recognition uses 3D-based networks. During inference, due to the addition of a temporal dimension, these networks are limited by memory and runtime to clips of a small spatial scale and a low number of frames. In \cite{carreira2017quo}, the authors use the whole video during inference, averaging predictions temporally. More recent studies that achieved state-of-the-art results processed numerous, but relatively short, clips during inference. In \cite{wang2018non}, inference is done by sampling ten clips evenly from the full-length video and average the softmax scores to achieve the final prediction. SlowFast \cite{feichtenhofer2019slowfast} follows the same practice and introduces the term \textit{“view”} -- a temporal clip with a spatial crop. SlowFast uses ten temporal clips with three spatial crops at inference time; thus, 30 different views are averaged for the final prediction. X3D \cite{feichtenhofer2020x3d} follows the same practice, but in addition, it uses larger spatial scales to achieve its best results on 30 different views.

This common practice of multi-view inference is somewhat counterintuitive, especially when handling long videos. A more intuitive way is to “look” at the entire video context before deciding on the action, rather than viewing only small portions of it. Fig.~\ref{fig:frames} shows 16 frames extracted evenly from a video of the \textit{abseiling} category. The actual action is obscured or not visible in several parts of the video; this might lead to a false action prediction in many views. The potential in focusing on the segments in the video that are most relevant is a powerful ability. However, full video inference produces poor performance in methods that were trained using short clips (Table~\ref{tb:6} and ~\ref{tb:7}). In addition, it is also limited in practice due to hardware, memory, and runtime aspects.


\section{Video Action Recognition with VTN}

In order to evaluate our approach and the impact of context attention on video action recognition, we use several spatial backbones pre-trained on 2D images. 

\paragraph{ViT-B-VTN.} Combining the state-of-the-art image classification model, ViT-Base \cite{dosovitskiy2021an}, as the backbone in VTN. We use a ViT-Base network that was pre-trained on ImageNet-21K. Using ViT as the backbone for VTN produces an end-to-end transformers-based network that uses attention both for the spatial and temporal domains.  

\paragraph{R50/101-VTN.} As a comparison, we also use a standard 2D ResNet-50 and ResNet-101 networks \cite{he2016deep}, pre-trained on ImageNet.

\paragraph{DeiT-B/BD/Ti-VTN.} Since ViT-Base was trained on ImageNet-21K we also want to compare VTN by using similar networks trained on ImageNet. We use the recent work of \cite{touvron2020training} and apply DeiT-Tiny, DeiT-Base, and DeiT-Base-Distilled as the backbone for VTN.

\begin{figure*}[t]
	\begin{center}
		\includegraphics[width=0.99\linewidth]{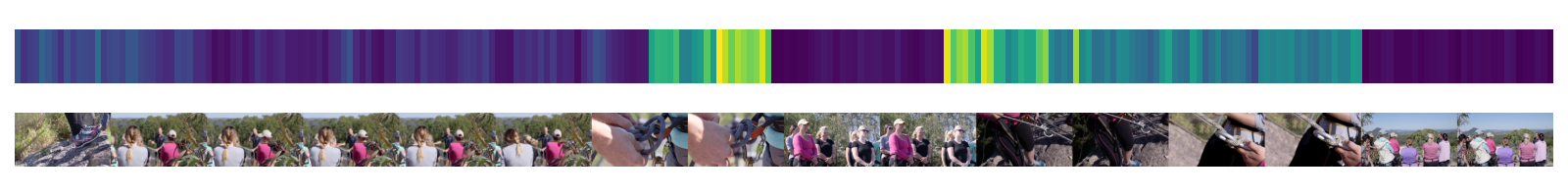}
	\end{center}
	\caption{Illustrating all the single-head first attention layer weights of the $[CLS]$ token \textit{vs.} 16 frames pulled evenly from a video. High weight values are represented by a \textit{warm} color (yellow) while low values by a \textit{cold} color (blue). The video’s segments in which \textit{abseiling} category properties are shown (\eg, shackle, rope) exhibit higher weight values compared to segments in which non-relevant information appears (\eg, shoes, people). The model prediction is \textit{abseiling} for this video.}
	\label{fig:colorbar}
\end{figure*}

\begin{figure}[t]
	\begin{center}
		\includegraphics[width=0.9\linewidth]{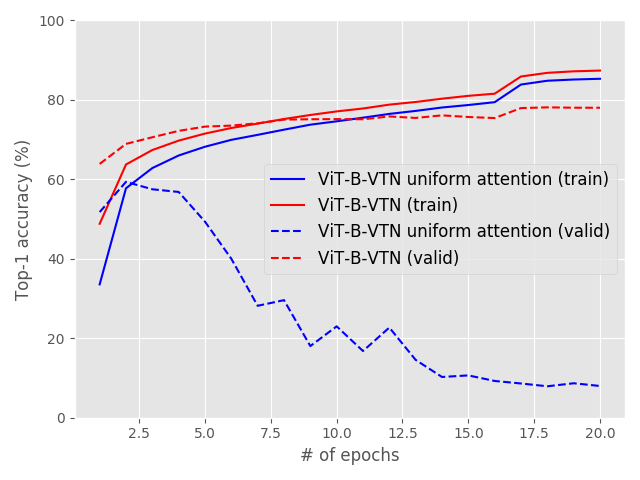}
	\end{center}
	\caption{Evaluating the influence of attention on the training (solid line) and validation (dashed line) curves for Kinetics-400. A similar ViT-B-VTN with three Longformer layers is trained for both cases, and we modify the attention heads between a learned one (red) and a fixed uniform version (blue).}
	\label{fig:attn_curves}
\end{figure}

\subsection{Implementation Details}

\paragraph{Training.} The spatial backbones we use were pre-trained on either ImageNet or ImageNet-21k. The Longformer and the MLP classification head were randomly initialized from a normal distribution with zero mean and 0.02 std. We train the model end-to-end using video clips. These clips are formed by choosing a random frame as the starting point, then sampling 2.56 or 5.12 seconds as the video’s temporal footprint. The final clip frames are subsampled uniformly to a fixed number of frames $N (N=16,32)$, depending on the setup. 

For the spatial domain, we randomly resize the shorter side of all the frames in the clip to a $[256,320]$ scale and randomly crop all frames to $224\times224$. Horizontal flip is also applied randomly on the entire clip.

The ablation experiments were done on a 4-GPU machine. Using a batch size of 16 for the ViT-VTN (on 16 frames per clip input) and a batch size of 32 for the R50/101-VTN. We use an SGD optimizer with an initial learning rate of $10^{-3}$ and a different learning rate reduction policy, steps-based for the ViT-VTN versions and cosine schedule decay for the R50/101-VTN versions. In order to report the wall runtime, we use an 8-V100-GPU machine.

Since we use 2D models as the spatial backbone, we can manipulate the input clip shape 
$x_\text{clip} \in \mathbb{R}^{B \times C \times T \times H \times W}$ 
by stacking all frames from all clips within a batch to create a single frames batch of shape 
$x \in \mathbb{R}^{ ( B \cdot T ) \times C \times H \times W  }$. 
Thus, during training, we propagate all batch frames in a single forward-backward pass.

For the Longformer, we use an \textit{effective attention window} of size 32, which was applied for each layer. Two other hyperparameters are the dimensions set for the \textit{Hidden size} and the \textit{FFN inner hidden size}. These are a direct derivative of the spatial backbone. Therefore, in R50/101-VTN we use 2048 and 4096, respectively, and for ViT-B-VTN we use 768 and 3072, respectively. In addition, we apply \textit{Attention Dropout} with a probability of 0.1. We also explore the impact of the number of Longformer layers.

The positional embedding (PE) information is only relevant for the temporal attention-based encoder (Fig.~\ref{fig:arch}). We explore three positional embedding approaches (Table~\ref{tb:3}): (1) \textit{Learned positional embedding} - since a clip is represented using frames taken from the full video sequence, we can learn an embedding that uses as input the frame location (index) in the original video, giving the Transformer information regarding the position of the clip in the entire sequence; (2) \textit{Fixed absolute encoding} - we use a similar method to the one in DETR \cite{carion2020end}, and modified it to work on the temporal axis only; and (3) \textit{No positional embedding} - no information is added in the temporal dimension, but we still use the global position to mark the special $[CLS]$ token position.

\begin{table*}
	\tablestyle{9.5pt}{1.05}
	\begin{center}
		\begin{tabular}{l|c|c|c|c|c|c|c}
			\multicolumn{1}{c|}{model} & \multicolumn{1}{c|}{\begin{tabular}[c]{@{}c@{}}training wall \\ runtime (minutes)\end{tabular}} & \multicolumn{1}{c|}{\begin{tabular}[c]{@{}c@{}}\# training \\ epochs\end{tabular}} & \multicolumn{1}{c|}{\begin{tabular}[c]{@{}c@{}}validation wall \\ runtime (minutes)\end{tabular}} & \multicolumn{1}{c|}{\begin{tabular}[c]{@{}c@{}}inference \\ approach\end{tabular}} & \multicolumn{1}{c|}{\begin{tabular}[c]{@{}c@{}}params \\ (M)\end{tabular}} & \multicolumn{1}{c|}{top-1} & \multicolumn{1}{c}{top-5} \\
			\hline
			I3D*                & 30                              & -                  & 84          & multi-view         & 28             & 73.5 \cite{fan2020pyslowfast}      & 90.8 \cite{fan2020pyslowfast}                 \\
			NL I3D (our impl.)       & 68                              & 50                 & 150              & multi-view         & 54                 & 74.1            & 91.7                      \\
			NL I3D (our impl.)       & 68                              & 50                 & 31                                & full-video         & 54                             & 72.1                      & 90.5                      \\
			SlowFast-8X8-R50*   & 70      & 196 \cite{feichtenhofer2019slowfast}    & 140          & multi-view  & 35    & 77.0 \cite{fan2020pyslowfast}   & 92.6 \cite{fan2020pyslowfast}        \\
			SlowFast-8X8-R50*   & 70           & 196 \cite{feichtenhofer2019slowfast}   & 26           & full-video         & 35                             & 68.4                      & 87.1                      \\
			SlowFast-16X8-R101* & 220  & 196 \cite{feichtenhofer2019slowfast}  & 244      & multi-view   & 60     & 78.9 \cite{fan2020pyslowfast}     & 93.5 \cite{fan2020pyslowfast}      \\
			\hline
			R50-VTN             & 62                              & 40                 & 32                                & full-video         & 168                            & 71.2                      & 90.0                      \\
			R101-VTN            & 110                             & 40                 & 32                                & full-video         & 187                            & 72.1                      & 90.3   \\
			DeiT-Ti-VTN (3 layers) & 52                              & 60                 & 30                                & full-video         & 10                             & 67.8                      & 87.5                      \\
			ViT-B-VTN (1 layer)    & 107                             & 25                 & 48                                & full-video         & 96                             & 78.6                      & 93.4                      \\
			ViT-B-VTN (3 layers)   & 130                             & 25                 & 52                                & full-video         & 114                            & 78.6                      & 93.7                 \\
			ViT-B-VTN (3 layers)$^\dagger$	  & 130                             & 35                 & 52                                & full-video         & 114                            & \textbf{79.8}                      & \textbf{94.2}
		\end{tabular}
	\end{center}
	\caption{To measure the overall time needed to train each model, we observe how long it takes to train a single epoch and how many epochs are required to achieve the best performance. We compare these numbers to the validation top-1 and top-5 accuracy on Kinetics-400 and the number of parameters per model. To measure the training wall runtime, we ran a single epoch for each model, on the same 8-V100-GPU machine, with a 16GB memory per GPU. The models marked by (*) were taken from the \textit{PySlowFast} GitHub repository \cite{fan2020pyslowfast}. We report the accuracy as written in the \textit{Model Zoo}, which was done using the 30 multi-view inference approach. To measure the wall runtime, we used the code base of \textit{PySlowFast}. To calculate the SlowFast-16X8-R101 time on the same GPU machine, we used a batch size of 16. The number of epochs is reported, when possible, based on the original model paper. All other models, including the NL I3D, are trained using our codebase and evaluated with a full video inference approach. ($\dagger$) The model in the last row was trained with extensive data augmentation.}
	\label{tb:6}
\end{table*}

\paragraph{Inference.} In order to show a comparison between different models, we use both the common practice of inference in multi-views and a full video inference approach (Sec.~\ref{section:looking}).

In the multi-view approach, we sample 10 clips evenly from the video. For each clip, we first resize the shorter side to 256, then take three crops of size $224 \times 224$ from the left, center, and right. The result is 30 views per video, and the final prediction is an average of all views' softmax scores. 

In the full video inference approach, we read all the frames in the video. Then, we align them for batching purposes, by either sub- or up-sampling, to 250 frames uniformly. In the spatial domain, we resize the shorter side to 256 and take a center crop of size $224 \times 224$.


\begin{figure}[t]
	\begin{center}
		\includegraphics[width=0.9\linewidth]{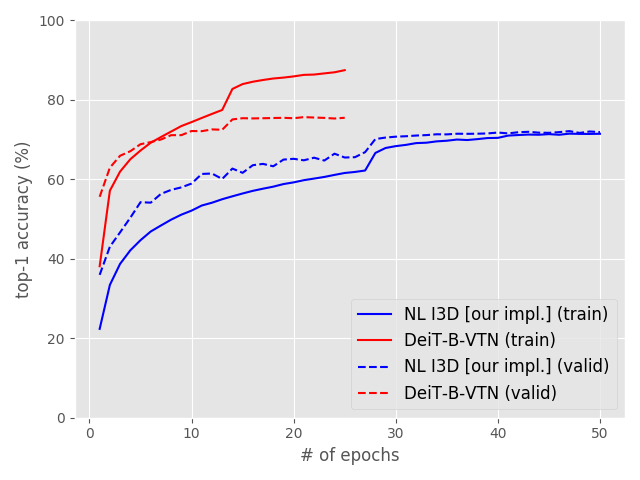}
	\end{center}
	\caption{Kinetics-400 learning curves for our implementation of NL I3D (blue) \textit{vs.} DeiT-B-VTN (red). We show the top-1 accuracy for the train set (solid line) and the validation set (dash line). Top-1 accuracy during training is calculated based on a single random clip, while during validation we use the full video inference approach. DeiT-B-VTN shows high performance in every step of the training and validation process. It reaches its best accuracy after only 25 epochs compared to the NL I3D that needs 50 epochs.}
	\label{fig:runtime}
\end{figure}

\section{Experiments}

\subsection{Ablation Experiments on Kinetics-400}

\paragraph{Kinetics-400 dataset.} The original Kinetics-400 dataset \cite{kay2017kinetics} consists of 246,535 training videos and 19,761 validation videos. Each video is labeled with one of 400 human action categories, curated from YouTube videos.
Since some YouTube links are expired, we could only download 234,584 of the original dataset, thus missing 11,951 videos from the training set, which are about 5\%. This leads to a slight drop in performance of about 0.5\%\footnote {\url{https://github.com/facebookresearch/video-nonlocal-net/blob/master/DATASET.md}}.

In the validation set, we are missing one video. To test our data’s validity and compare it to previous studies, we evaluated the SlowFast-8X8-R50 model, published in \textit{PySlowFast}\cite{fan2020pyslowfast}, on our validation data. We got 76.45\% top1-accuracy \textit{vs.} the reported 77\%, thus a drop of 0.55\%. This drop might be related to different \textit{FFmpeg} encoding and rescaling of the videos. From this point forward, when comparing to other networks, we report results taken from the original studies except when we evaluate them on the full video inference in which we use our validation set. All our approach results are reported based on our validation set.

\paragraph{Spatial backbone variations.} We start by examining how different spatial backbone architectures impact VTN performance. Table~\ref{tb:1} shows a comparison of different VTN variants and the pretrain dataset the backbone was first trained on. ViT-B-VTN is the best performing model and reaches 78.6\% top-1 accuracy and 93.7\% top-5 accuracy. The pretraining dataset is important. Using the same ViT backbone, only changing between DeiT (pre-trained on ImageNet) and ViT (pre-trained on ImageNet-21K) we get an improvement in the results.

\begin{table*}
	\tablestyle{12pt}{1.05}
	\begin{center}
		\begin{tabular}{l|c|c|c|c|c|c}
			\multicolumn{1}{c|}{model} & \multicolumn{1}{c|}{\begin{tabular}[c]{@{}c@{}}inference \\ approach\end{tabular}}  & \multicolumn{1}{c|}{\begin{tabular}[c]{@{}c@{}}\# frames per \\ inference view\end{tabular}} & \multicolumn{1}{c|}{\# of views} & \multicolumn{1}{c|}{\begin{tabular}[c]{@{}c@{}}test crop \\ size\end{tabular}} & \multicolumn{1}{c|}{\begin{tabular}[c]{@{}c@{}}inference \\ GFLOPs\end{tabular}} & top-1     \\
			\hline
			NL I3D (our impl.)       & multi-view       & 32                 & 30                               & 224                        & 2,625                                  & 74.1                             \\
			NL I3D (our impl.)        & full-video      & 250                                & 1                & 224                                  & 1,266                                          & 72.2                             \\
			\hline
			SlowFast-8X8-R50*      & multi-view    & 32                   & 30                      & 256                & 1,971                                  & 77.0  \cite{fan2020pyslowfast}                          \\
			SlowFast-8X8-R50*      & full-video    & 250                 & 1                                & 256                         & 517                                 & 68.4                             \\
			SlowFast-16X8-R101*     & multi-view   & 64                                   & 30                               & 256                       & 6,390              & 78.9 \cite{fan2020pyslowfast}                        \\
			SlowFast-16X8-R101*     & full-video   & 250         & 1                                & 256                  & 838                                                                              & -                                \\
			\hline
			R50-VTN (3 layers)      & multi-view   & 16              & 30                               & 224                                    & 2,106                                 & 70.9                             \\
			R50-VTN (3 layers)      & full-video   & 250                       & 1         & 224            & 1,059                                                                            & 71.2                             \\
			R101-VTN (3 layers)    & multi-view    & 16                          & 30                               & 224                              & 3,895                                 & 72.0                             \\
			R101-VTN (3 layers)     & full-video  & 250                        & 1                                & 224                 & 1,989                                        & 72.1                             \\
			ViT-B-VTN (1 layer)    & multi-view    & 16                         & 30                                & 224                          & 8,095                            & 78.5                             \\
			ViT-B-VTN (1 layer)     & full-video   & 250                            & 1                                & 224                    & 4,214                                & 78.6                             \\
			ViT-B-VTN (3 layers) & multi-view      & 16                                 & 30                                & 224                           & 8,113                                             & 78.6                             \\
			ViT-B-VTN (3 layers)    & full-video   & 250                             & 1                                & 224                             & 4,218                           & 78.6    
		\end{tabular}
	\end{center}
	\caption{Comparing the number of GFLOPs during inference. The models marked by (*) were taken from the \textit{PySlowFast} GitHub repository \cite{fan2020pyslowfast}. We reproduced the SlowFast-8X8-R50 results by using the repository and our Kinetics-400 validation set and got 76.45\% compared to the reported value of 77\%. When running this model using a full video inference approach, we get a significant drop in performance of about 8\%. We did not run the SlowFast-16X8-R101 because it was not published. The inference GFLOPs is reported by multiplying the number of views with the GFLOPs calculated per view. ViT-B-VTN with one layer achieves 78.6\% top-1 accuracy, a 0.3\% drop compared to SlowFast-16X8-R101 while using $1.5\times$ fewer GFLOPS.}
	\label{tb:7}
\end{table*}


\paragraph{Longformer depth.} Next, we explore how the number of attention layers impacts the performance. Each layer has 12 attention heads and the backbone is ViT-B. Table~\ref{tb:2} shows the validation top-1 and top-5 accuracy for 1, 3, 6, and 12 attention layers. The comparison shows that the difference in performance is small. This is counterintuitive to the fact that deeper is better. It might be related to the fact that Kinetics-400 videos are relatively short, around 10 seconds. We believe that processing longer videos will benefit from a large receptive field obtained by using a deeper Longformer.


\paragraph{Longformer positional embedding.} In Table~\ref{tb:3} we compare three different positional embedding methods, focusing on learned, fixed, and no positional embedding. All versions are done with a ViT-B-VTN, a temporal footprint of 5.12 seconds, and a clip size of 16 frames. Surprisingly, the one without any positional embedding achieved slightly better results than the fixed and learned versions. 

As this is an interesting result, we also use the same trained models and evaluate them after randomly shuffling the input frames only in the validation set videos. This is done by first taking the unshuffled frame embeddings, then shuffle their order, and finally add the positional embedding. This raised another surprising finding, in which the shuffle version gives better results, reaching 78.9\% top-1 accuracy on the no positional embedding version. Even in the case of learned embeddings it does not have a diminishing effect. Similar to the Longformer depth, we believe that this might be related to the relatively short videos in Kinetics-400, and longer sequences might benefit more from positional information. We also argue that this could mean that Kinetics-400 is primarily a static frame, appearance based classification problem rather than a motion problem \cite{sevilla2021only}.


\paragraph{Temporal footprint and number of frames in a clip.} We also explore the effect of using longer clips in the temporal domain and compare a temporal footprint of 2.56 \textit{vs.} 5.12 seconds. And also how the number of frames in the clip impact the network performance. The comparison is done on a ViT-B-VTN with one attention layer in the Longformer. Table~\ref{tb:4} shows that top-1 and top-5 accuracy are similar, implying that VTN is agnostic to these hyperparameters.


\paragraph{Finetune the 2D spatial backbone.} Instead of fine-tuning the spatial backbone, by continuing the back-propagation process, when training VTN, we can use a \textit{frozen} 2D network solely for feature extraction. Table~\ref{tb:5} shows the validation accuracy when training a ViT-B-VTN with three attention layers with and without also training the backbone. Fine-tuning the backbone improves the results by 7\% in Kinetics-400 top-1 accuracy.


\begin{table*}
	\tablestyle{12pt}{1.05}
	\begin{center}
		\begin{tabular}{l|c|c|c|c|c}
			\multicolumn{1}{c|}{model}                                & pretrain    & \begin{tabular}[c]{@{}c@{}}MiT version\end{tabular} & inputs   & top-1                            & top-5                            \\ \hline
			ResNet50                            & ImageNet    & v1                                                    & RGB      & 27.2 \cite{monfort2019moments}         & 51.7 \cite{monfort2019moments}         \\
			I3D                                 & ImageNet    & v1                                                    & RGB + OF & 29.5 \cite{monfort2019moments}         & 56.1 \cite{monfort2019moments}         \\
			AssembelNet-50 \cite{Ryoo2020AssembleNet}  & Kinetics400 & v1                                                    & RGB + OF & 33.9                            & 60.9                            \\
			AssembelNet-101 \cite{Ryoo2020AssembleNet} & Kinetics400 & v1                                                    & RGB + OF & 34.3                            & 62.7                            \\ \hline
			I3D                                 & ImageNet    & v2                                                    & RGB      & 28.4*  & 54.5* \\
			NL I3D (our impl.)                  & Kinetics400 & v2      & RGB      & 30.1                       & 57.3                             \\ \hline
			ViT-B-VTN                           & Kinetics400 & v2                                                    & RGB      & \textbf{37.4}                             & 65.3                             \\
			ViT-B-VTN (w/ shuffle)              & Kinetics400 & v2                                                    & RGB      & \textbf{37.4}                   & \textbf{65.4}                   
		\end{tabular}
	\end{center}
	\caption[]{Comparison with the state-of-the-art on MiT-v1 and MiT-v2. The results marked by (*) are based on MiT GitHub repository\footnotemark.}
	\label{tb:8}
\end{table*}

\paragraph{Does attention matter?} A key component in our approach is the impact of attention functionally on the way VTN perceives the full video sequence. To convey this impact we train two VTN networks, using three layers in the Longformer, but with a single head for each layer. In one network the head is trained as usual, while in the second network instead of computing attention based on query/key dot products and softmax, we replace the attention matrix with a hard-coded uniform distribution that is not updated during back-propagation. 

Fig.~\ref{fig:attn_curves} shows the learning curves of these two networks. Although the training has a similar trend, the learned attention performs better. In contrast, the validation of the uniform attention collapses after a few epochs demonstrating poor generalization of that network. Further, we visualize the $[CLS]$ token attention weights by processing the same video from Fig.~\ref{fig:frames} with the single-head trained network and depicted, in Fig.~\ref{fig:colorbar}, all the weights of the first attention layer aligned to the video’s frames. Interestingly, the weights are much higher in segments related to the \textit{abseiling} category. In Appendix~\ref{sec:AppendixA}. we show a few more examples.



\paragraph{Training and validation runtime.} An interesting observation we make concerns the training and validation wall runtime of our approach. Although our networks have more parameters, and therefore, are longer to train and test, they are actually much faster to converge and reach their best performance earlier. Since they are evaluated using a single view of all video frames, they are also faster during validation. Table~\ref{tb:6} shows a comparison of different models and several VTN variants. Compared to the state-of-the-art SlowFast model, our ViT-B-VTN with one layer achieves almost the same results but completes an epoch faster while requiring fewer epochs. This accumulates to a $16.1\times$ faster end-to-end training. The validation wall runtime is also $5.1\times$ faster due to the full video inference approach.

To better demonstrate the fast convergence of our approach, we wanted to show an apples-to-apples comparison of different training and evaluating curves for various models. However, since other methods use the multi-view inference only post-training, but use a single view evaluation while training their models, this was hard to achieve. Thus, to show such comparison and give the reader additional visual information, we trained a NL I3D (pre-trained on ImageNet) with a full video inference protocol during validation (using our codebase and reproduced the original model results). We compare it to DeiT-B-VTN which was also pre-trained on ImageNet. Fig.~\ref{fig:runtime} shows that the VTN-based network converges to better results much faster than the NL I3D and enables a much faster training process compared to 3D-based networks.



\paragraph{Data augmentation.} Recent studies showed that data augmentation significantly improves the performance of transformers-based models \cite{touvron2020training}. To demonstrate its impact on VTN, we apply extensive data augmentation as suggested in DeiT \cite{touvron2020training} and RandAugment \cite{cubuk2020randaugment}. Table~\ref{tb:6} shows that our method reaches 79.8\% top-1 accuracy, a 1.2\% improvement \textit{vs.} the same model trained without such augmentations. Training with augmentations requires 10 more epochs but didn't impact the training wall runtime.

\paragraph{Final inference computational complexity.} Finally, we examine what is the final inference computational complexity for various models by measuring GFLOPs. Although other models need to evaluate multiple views to reach their highest performance, ViT-B-VTN performs almost the same for both inference protocols. Table~\ref{tb:7} shows a significant drop of about 8\% when evaluating the SlowFast-8X8-R50 model using the full video approach. In contrast, ViT-B-VTN maintains the same performance while requiring, end-to-end, fewer GFLOPs at inference. 



\subsection{Experiments on Moments in Time}

The Moments in Time (MiT) dataset is a large-scale collection of short (3 seconds) videos \cite{monfort2019moments}. MiT is a challenging dataset, with state-of-the-art results just above 34\% top-1 accuracy \cite{Ryoo2020AssembleNet}. In this work, we use MiT-v2, consisting of 727,305 training videos and 30,500 validation videos. Each video is labeled with one of 305 classes of dynamic events. Although previous studies worked on MiT-v1 (802,264 training videos, 33,900 validation videos, 339 classes), this dataset is no longer available. In Table~\ref{tb:8}, we show the results of various models on MiT-v1 and MiT-v2. Since the relation between v1 and v2 in terms of performance was not established and thus unknown, we also trained our implementation of NL I3D on MiT-v2 using RGB inputs and achieved comparable results to those of I3D (RGB+OF) published on MiT-v1 \cite{monfort2019moments}. Furthermore, ViT-B-VTN achieves the highest top-1 accuracy on MiT-v2 while using only RGB frames as input.


\footnotetext{\url{https://github.com/zhoubolei/moments_models}}

\section{Conclusion}

We presented a modular transformer-based framework for video recognition tasks. Our approach introduces an efficient way to evaluate videos at scale, both in terms of computational resources and wall runtime. It allows full video processing during test time, making it more suitable for dealing with long videos. Although current video classification benchmarks are not ideal for testing long-term video processing ability, hopefully, in the future, when such datasets become available, models like VTN will show even larger improvements compared to 3D ConvNets.

\paragraph{Acknowledgements.} We thank Ross Girshick for providing valuable feedback on this manuscript and for helpful suggestions on several experiments.


{\small
	\bibliographystyle{ieee_fullname} 
	\bibliography{vtn_bib}
}


\appendix

\begin{figure*}[h]
\section*{Appendix}
\section{Additional Qualitative Results}
\label{sec:AppendixA}
\begin{center}
\subfloat[][Label: \textit{Tai chi}. Prediction: \textit{Tai chi}.]{\includegraphics[width=0.95\linewidth]{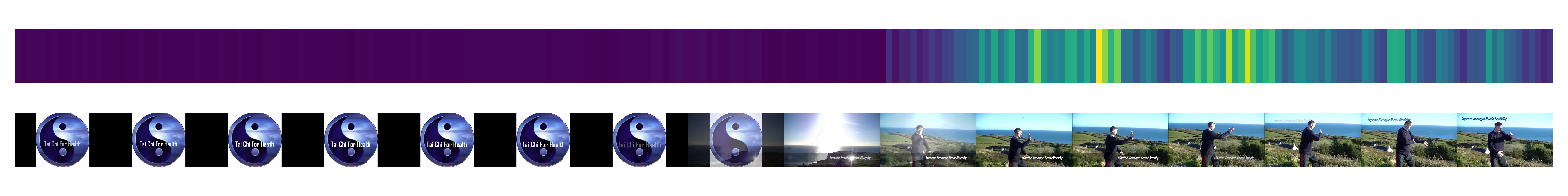}}

\subfloat[][Label: \textit{Chopping wood}. Prediction: \textit{Chopping wood}.]{\includegraphics[width=0.95\linewidth]{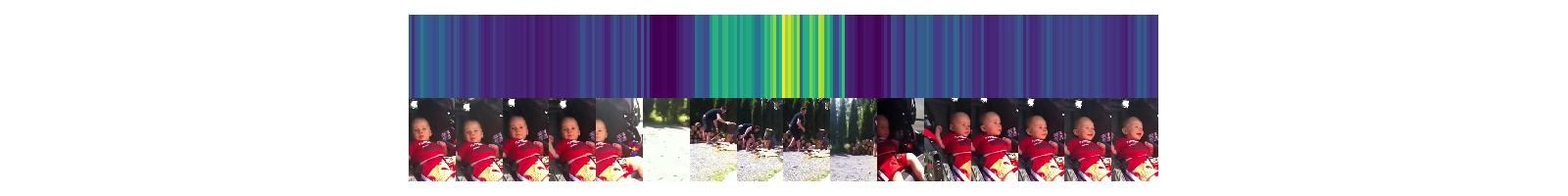}}

\subfloat[][Label: \textit{Archery}. Prediction: \textit{Archery}.]{\includegraphics[width=0.95\linewidth]{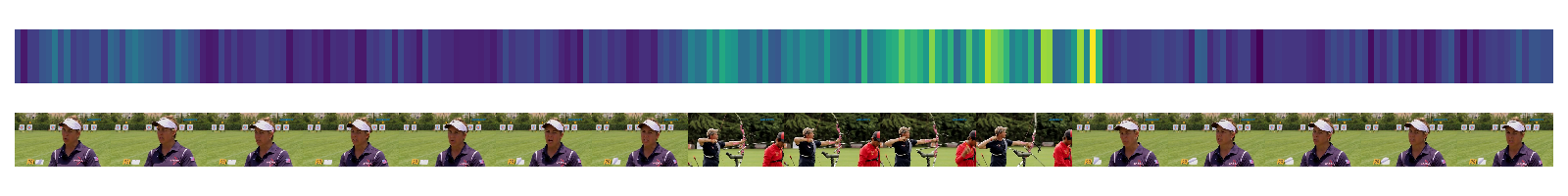}}

\subfloat[][Label: \textit{Throwing discus}. Prediction: \textit{Flying kite}.]{\includegraphics[width=0.95\linewidth]{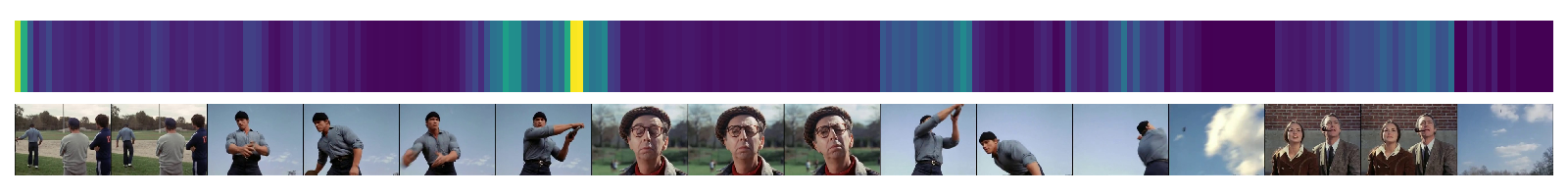}}

\subfloat[][Label: \textit{Surfing water}. Prediction: \textit{Parasailing}.]{\includegraphics[width=0.95\linewidth]{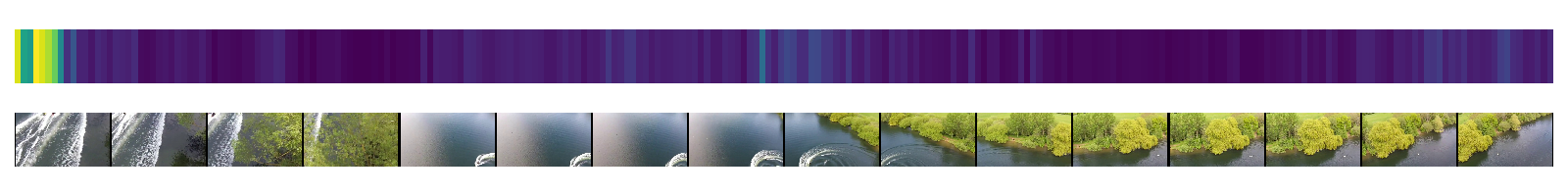}}
\end{center}
\caption{Additional qualitative examples of why attention matters. Similar to Fig.~\ref{fig:colorbar}, we illustrate the $[CLS]$ token weights of the first attention layer. We show some examples for successful predictions (a, b, c) and some of the failure modes of our approach (d, e).}
\label{fig:colorbar-appendix}
\end{figure*}


\end{document}